\documentclass{article}

\usepackage{microtype}
\usepackage{graphicx}
\usepackage{subfigure}
\usepackage{booktabs} 
\usepackage{tikz}
\usepackage{pgfplots}

\usepackage[square,numbers,sort&compress]{natbib}
\usepackage{enumitem}
\usepackage{multirow}
\usepackage{booktabs}
\usepackage{caption}
\usepackage[normalem]{ulem}
\usepackage{xparse}
\usepackage[font={small}]{caption}
\usepackage{amsmath}  
\usepackage{amsfonts}
\usepackage{bm}
\usepackage{colortbl}

\usepackage{tabularx}
\newcolumntype{Y}{>{\centering\arraybackslash}X}

\newcommand{\newterm}[1]{{\bf #1}}

\def\eqref#1{equation~\ref{#1}}

\def\1{\bm{1}}

\def\vtheta{{\bm{\theta}}}

\def\vx{{\bm{x}}}

\def\vz{{\bm{z}}}

\DeclareMathAlphabet{\mathsfit}{\encodingdefault}{\sfdefault}{m}{sl}
\SetMathAlphabet{\mathsfit}{bold}{\encodingdefault}{\sfdefault}{bx}{n}

\def\sM{{\mathbb{M}}}

\newcommand{\ptrain}{\hat{p}_{\rm{data}}}

\newcommand{\E}{\mathbb{E}}

\newcommand{\normlone}{L^1}
\newcommand{\normltwo}{L^2}

\newcommand{\normmax}{L^\infty}

\DeclareMathOperator*{\argmin}{arg\,min}

\definecolor{Gray}{gray}{0.85}
\definecolor{LightCyan}{rgb}{0.88,1,1}

\usepackage{hyperref}

\usepackage[arxiv]{icml2018}

\icmltitlerunning{Adversarial Logit Pairing}

\begin{document}

\twocolumn[
\icmltitle{Adversarial Logit Pairing}



\icmlsetsymbol{equal}{*}

\begin{icmlauthorlist}
\icmlauthor{Harini Kannan*}{goo}
\icmlauthor{Alexey Kurakin}{goo}
\icmlauthor{Ian Goodfellow}{goo}
\end{icmlauthorlist}

\icmlaffiliation{goo}{Google Brain, Mountain View, CA, USA}

\icmlcorrespondingauthor{Harini Kannan}{hkannan@google.com}
\icmlcorrespondingauthor{Alexey Kurakin}{kurakin@google.com}
\icmlcorrespondingauthor{Ian Goodfellow}{goodfellow@google.com}

\icmlkeywords{Machine Learning, ICML}

\vskip 0.3in
]



\printAffiliationsAndNotice{}  

\newcommand{\etal}{\textit{et al}. }

\begin{abstract}

In this paper, we develop improved techniques for defending against adversarial examples at scale.
First, we implement the state of the art version of adversarial training
at unprecedented scale on ImageNet and investigate whether it remains effective in this setting---an
important open scientific question \cite{obfuscated}.
Next, we introduce enhanced defenses using a technique
we call logit pairing, a method that encourages logits for pairs of examples
to be similar.
When applied to clean examples and their adversarial counterparts, logit pairing improves accuracy on adversarial examples over vanilla adversarial training; we also find that logit pairing on clean examples only is competitive with adversarial training in terms of accuracy on two datasets. 
Finally, we show that adversarial logit pairing achieves the \textbf{state of the art} defense on Imagenet against PGD white box attacks, with an accuracy improvement from \textbf{1.5\% to 27.9\%}. Adversarial logit pairing also successfully damages the current state of the art defense against
black box attacks on Imagenet~\cite{eat}, dropping its accuracy from \textbf{66.6\% to 47.1\%}.
With this new accuracy drop, adversarial logit pairing ties with \citet{eat} for the state of the art on black box attacks on ImageNet.

\end{abstract}

\section{Introduction}

Many deep learning models today are vulnerable to \newterm{adversarial examples},
or inputs that have been intentionally optimized to cause misclassification.
In the context of computer vision, object recognition classifiers incorrectly
recognize images that have been modified with small, often imperceptible
perturbations.
It is important to develop models that are robust to adversarial perturbations
for a variety of reasons:  
\begin{itemize}
    \item so that machine learning can be used in situations
where an attacker may attempt to interfere with the operation of the deployed
system,
    \item so that machine learning is more useful for model-based optimization,
    \item to gain a better understanding of how to provide performance guarantees for
models under distribution shift,
\item to gain a better understanding of how to
enforce smoothness assumptions, etc.
\end{itemize}

In this paper, we investigate defenses against such adversarial attacks. The contributions of this paper are the following:

\begin{itemize}
  \item We implement the state of the art version of adversarial training at unprecedented scale and investigate its effectiveness on the ImageNet dataset.
  
  \item We propose \newterm{logit pairing}, a method that encourages the logits for two pairs of examples to be similar. We propose two flavors of logit pairing: clean and adversarial.
  \item We show that \newterm{clean logit pairing} is a method with minimal computational cost that defends against PGD black box attacks almost as well as adversarial training for two datasets. 
  \item We show that \newterm{adversarial logit pairing} is a method that leads to higher accuracy when subjected to white box and black box attacks. We achieve the current state of the art on black-box and white-box accuracies with our model trained with adversarial logit pairing.
  \item We show that attacks constructed with our adversarially trained models \textbf{substantially damage} the current state of the art for black box defenses on ImageNet \cite{eat}. We then show that our models are resistant to these attacks.
 
\end{itemize}

\section{Definitions and threat models}

Defense mechanisms are intended to provide security under particular threat models.
The threat model specifies the capabilities of the adversary.
In this paper, we always assume the adversary is capable of forming attacks
that consist of perturbations of limited $\normmax$
norm. This is a simplified task chosen because it is more amenable to benchmark evaluations. Realistic attackers against computer vision systems would likely use different attacks that are difficult to characterize with norm balls, such as \citet{brown2017adversarial}.
We consider two different threat models characterizing amounts of information the adversary can have:
\begin{enumerate}
    \item \newterm{White box}: the attacker has full information about the model (i.e. knows the architecture, parameters, etc.).
    \item \newterm{Black box}: the attacker has no information about the model's architecture or parameters, and no ability to send queries to the model to gather more information. 
\end{enumerate}

\section{The challenges of defending ImageNet classifiers}

Multiple methods to defend against adversarial examples have been proposed \citep{szegedy2013intriguing, goodfellow2014explaining, papernot2016distillation, xu2017feature, eat, madry, kolter2017provable, buckman2018thermometer}.
Recently, \citet{obfuscated} broke several defenses proposed for the white box setting that relied
on empirical testing to establish their level of robustness.
In our work we choose to focus on \citet{madry} because it is a method
that has withstood intense scrutiny even in the white box setting.
\citet{obfuscated} endorsed \citet{madry} as the only such method that they
were not able to break.
However, they observe that the defense from \citet{madry} has not been shown to scale to ImageNet.
There are also certified defenses \citep{kolter2017provable,raghunathan2018certified,sinha2018certifiable} that provide 
guaranteed robustness, but the total amount of robustness they guarantee is small
compared to the amount empirically claimed by \citet{madry}.
This leaves \citet{madry} as a compelling defense to study because it provides a large benefit that has withstood
intensive scrutiny.

In this paper, we implement the \citet{madry} defense at ImageNet scale for the first time
and evaluate it using the same
attack methodology as has been used at smaller scale.
Our results provide an important conclusive answer to an open question \cite{obfuscated} about whether this defense strategy scales.

The defense used by \citet{madry} consists of using adversarial training \citep{szegedy2013intriguing, goodfellow2014explaining} with an attack called ``projected gradient descent'' (PGD).
Their PGD attack consists of initializing the search for an adversarial example at a random point
within the allowed norm ball, then running several iterations of the basic iterative method \citep{kurakin17physical}
to find an adversarial example. The noisy initial point creates a stronger attack than other previous iterative methods such as BIM \cite{kurakin17}, and performing adversarial training with this stronger attack makes their defense more successful \cite{madry}.
\citet{kurakin17} earlier reported that adversarial training with (non-noisy) BIM adversarial examples
did not result in general robustness to a wide variety of attacks.

All previous attempted defenses on ImageNet \cite{kurakin17,eat} report \textbf{error rates of 99 percent} on strong, multi-step white box attacks.
We, for the first time, scale the \citet{madry} defense to this setting and successfully
apply it. Furthermore, we also introduce an enhanced defense that greatly improves over this baseline and improves the amount of robustness achieved.

\section{Methods}

\subsection{Adversarial training}

\citet{madry} suggests that PGD is a universal first order adversary -- in other words, developing robustness against PGD attacks also implies resistance against many other first order attacks. We use adversarial training with PGD
as the underlying basis for our methods:

\begin{equation}\label{eq:madry_training}
\argmin_{\theta}
\E_{(x, y) \in
\ptrain}
\Bigl( 
\max_{\delta \in S}
L(\theta, x + \delta, y)
\Bigr)
\end{equation}
where $\ptrain$ is the underlying training data distribution,
$L(\theta, x, y)$ is a loss function at data point $x$ which has true class $y$ for a model with parameters $\theta$,
and the maximization with respect to $\delta$ is approximated using noisy BIM.

We find that we achieve better performance not by literally solving the min-max problem
described by \citet{madry}. Instead, we train on a mixture of clean and adversarial examples,
as recommended by \citet{goodfellow2014explaining,kurakin17}:

\begin{multline}\label{eq:mixed_adv_training}
\argmin_{\theta} \bigg[ \E_{(x, y) \in \ptrain} \Big( \max_{\delta \in S} L(\theta, x + \delta, y) \Big) +\\
\E_{(x, y) \in \ptrain} \Big( L(\theta, x, y) \Big) \bigg]
\end{multline}

This formulation helps to maintain good accuracy on clean examples.
We call this defense formulation \newterm{mixed-minibatch PGD} (M-PGD).
We note that though we have varied the defense
slightly from the one used in \citet{madry},
we still use the attack from \citet{madry},
which is also endorsed by \citet{obfuscated}.

\subsection{Logit pairing}

We propose \newterm{logit pairing}, a method to encourage the logits from two images to be similar to each other.
For a model that takes inputs $\vx$ and computes a vector of logits $\vz = f(\vx)$, logit pairing adds a loss
\[ \lambda L\left(f(\vx), f(\vx')\right) \]
for pairs of training examples $\vx$ and $\vx'$, where $\lambda$ is a coefficient determining the strength of the logit pairing penalty
and $L$ is a loss function encouraging the logits to be similar. In this paper we use $\normltwo$ loss
for $L$, but other losses such as $\normlone$ or Huber could also be suitable choices.

We explored two logit pairing techniques which are described below.
We found each of them to be useful:
adversarial logit pairing obtains the best-yet defense against the
Madry attack, while clean logit pairing, and
a related idea we call logit squeezing, provide
competitive defenses at significantly reduced cost.

\subsubsection{Adversarial logit pairing}
\newterm{Adversarial logit pairing} (ALP) matches the logits from a clean image
$\vx$ and its corresponding adversarial image $\vx'$.
In traditional adversarial training, the model is trained to assign
both $\vx$ and $\vx'$ to the same output class label, but the model
does not receive any information indicating that $\vx'$ is more similar
to $\vx$ than to another example of the same class.
ALP provides an extra regularization term encouraging similar embeddings
of the clean and adversarial versions of the same example, helping guide the
model towards better internal representations of the data.

Consider a model with parameters $\vtheta$
trained on a minibatch $\sM$ of clean examples $\{\vx^{(1)}, \dots, \vx^{(m)}\}$
and corresponding adversarial examples $\{\tilde{\vx}^{(1)}, \dots, \tilde{\vx}^{(m)}\}$.
Let $f(\vx; \vtheta)$ be the function mapping from inputs to logits of the model.
Let $J(\sM, \vtheta)$ be the cost function used for adversarial training
(the cross-entropy loss applied to train the classifier on each example
in the minibatch, plus any weight decay, etc.).
Adversarial logit pairing consists of minimizing the loss
\[
J(\sM, \vtheta) + \lambda \frac{1}{m} \sum_{i=1}^m L\left(f(\vx^{(i)}; \vtheta),
f(\tilde{\vx}^{(i)}; \vtheta) \right).
\]

\subsubsection{Clean logit pairing}
In \newterm{clean logit pairing} (CLP), $\vx$ and $\vx'$ are two randomly selected
{\em clean} training examples, and thus are typically not even from the same
class.
Let $J^{(\text{clean})}(\sM, \vtheta)$ be the loss function
used to train a classifier on a minibatch $\sM$,
such as a cross-entropy loss and any other loss terms
such as weight decay.
Clean logit pairing consists of minimizing the loss
\[
J^{(\text{clean})}(\sM, \vtheta) + \lambda \frac{2}{m} \sum_{i=1}^{\frac{m}{2}} L\left(f(\vx^{(i)}; \vtheta),
f(\vx^{(i+\frac{m}{2})}; \vtheta) \right).
\]

We included experiments with clean logit pairing in order to perform an
ablation study, understanding the contribution of the pairing loss 
itself relative to the formation of clean and adversarial pairs.
To our surprise, inducing similarity between random pairs of logits led to high levels of robustness on MNIST and SVHN.
This leads us to suggest clean logit pairing
as a method worthy of study in its own right
rather than just as a baseline.
CLP is surprisingly effective and has significantly lower computation cost
than adversarial training or ALP.

We note that our best results with CLP relied on adding Gaussian noise to the
input during training, a standard neural network regularization technique \citep{SietsmaDow91}.

\subsubsection{Clean logit squeezing}

Since clean logit pairing led to high accuracies, we hypothesized that the model was learning to predict logits of smaller magnitude and therefore being penalized for becoming overconfident. To this end, we tested penalizing the norm of the logits, which we refer to as ``logit squeezing'' for the rest of the paper. For MNIST, it turned out that logit squeezing gave us better results than logit pairing.

\section{Adversarial logit pairing results and discussion}

 \subsection{Results on MNIST}
 Here, we first present results with adversarial logit pairing on MNIST. We found that the exact value of the logit pairing weight did not matter too much on MNIST as long as it was roughly between 0.2 and 1. As long as \textit{some} logit pairing was added, the accuracy on adversarial examples improved compared to vanilla adversarial training. We used a final logit pairing weight of 1 in the values reported in Table \ref{fig:mnist_alp}. A weight of 1 corresponds to weighting both the adversarial logit pairing loss and the cross-entropy loss equally.
 
 We used the LeNet model as in \citet{madry}. We also used the same attack parameters they used: total adversarial perturbation of 76.5/255 (0.3), perturbation per step of 2.55/255 (0.01), and 40 total attack steps with 1 random restart. Similar to \citet{madry}, we generated black box examples for MNIST by independently initializing and adversarially training a copy of the LeNet model. We then used the PGD attack on this model to generate the black box examples.
 
\begin{table}[h]
\centering
\begin{tabular}{*{4}{c}}

\toprule
\rowcolor{Gray}
\textbf{Method} & \textbf{White Box} & \textbf{Black Box} & \textbf{Clean} \\
\midrule
M-PGD & 93.2\% & 96.0\% & 98.5\% \\
ALP & \textbf{96.4\%} & \textbf{97.5\%} & \textbf{98.8\%} \\
\bottomrule
\end{tabular}
  \caption{Comparison of adversarial logit pairing and vanilla adversarial training on MNIST. All accuracies reported are for the PGD attack.}
\label{fig:mnist_alp}
\end{table}
 
 As shown in Table \ref{fig:mnist_alp}, adversarial logit pairing achieves state of the art on MNIST for the PGD attack. It improves white box accuracy from 93.2\% to \textbf{96.4\%}, and it improves black box accuracy from 96.0\% to \textbf{97.5\%}.

\subsection{Results on SVHN}

\begin{table}[h]
\centering
\begin{tabular}{*{5}{c}}
\toprule
\rowcolor{Gray}
\textbf{Method} & \textbf{White Box} & \textbf{Black Box} & \textbf{Clean} \\
\midrule
M-PGD & 44.4\% & 55.4\% & \textbf{96.9\%} \\
ALP & \textbf{46.9\%} & \textbf{56.2\%} & 96.2\% \\
\bottomrule
\end{tabular}
  \caption{Comparison of adversarial logit pairing and vanilla adversarial training on SVHN. All accuracies reported are for the PGD attack.}
\label{fig:svhn_alp}
\end{table}

Our PGD attack parameters for SVHN were as follows: a total epsilon perturbation of 12/255, a per-step epsilon of 3/255, and 10 attack iterations.

For SVHN, we used the RevNet-9 model \cite{revnets}. RevNets are similar to ResNets in that they both use residual connections, have similar architectures, and get similar accuracies on multiple datasets. However, RevNets have large memory savings compared to ResNets, as their memory usage is constant and does not scale with the number of layers. Because of this, we used RevNets in order to take advantage of larger batch sizes and quicker convergence times. 

Similar to MNIST, most logit pairing values from 0.5 to 1 worked, and as long as some logit pairing was added, it greatly
improved accuracies. However, making the logit pairing values too large (e.g. anything larger than 2) did not lead to any benefit and was roughly the same as vanilla adversarial training. The final adversarial logit pairing weight used in Table \ref{fig:svhn_alp} was 0.5.

\subsection{Results on ImageNet}
\subsubsection{Motivation}
Prior to this work, the standard baseline of PGD adversarial training had not yet been scaled to ImageNet. Kurakin \etal \yrcite{kurakin17} showed that adversarial training with one-step attacks confers robustness to other one-step attacks, but is unable to make a difference with multi-step attacks. Training on multi-step attacks did not help either.
\citet{madry} demonstrated successful defenses based on multi-step {\em noisy} PGD adversarial training on MNIST and CIFAR-10, but did not scale the process to ImageNet. 

 Here, we implement and scale the state of the art
 adversarial training method from CIFAR-10 and MNIST to ImageNet for the first time. We then implement our adversarial logit pairing method for comparison. 
\subsubsection{Implementation details}
The cost of adversarial training scales with the number of attack steps because a full round of backpropagation is performed with each step. This means that a rough estimate of the total adversarial training time of a model can be found by multiplying the total clean training time by the number of attack steps. With ImageNet, this can be especially costly without any optimizations.

To effectively scale up adversarial training with PGD to ImageNet, we implemented synchronous distributed training in Tensorflow with 53 workers: 50 were used for gradient aggregation, and 3 were left as backup replicas. Each worker had one p100 card. We experimented with asynchronous gradient updates, but we found that it led to stale gradients and poor convergence. Additionally, we used 17 parameter servers that ran on CPUs. Large batch training helped to scale up adversarial training as well: each replica had a batch size of 32, for an effective batch size of 1600 images. We found that the total time to convergence was approximately 6 days.

Similar to \citet{kurakin17}, we use the InceptionV3 model to implement adversarial training on ImageNet in order to better compare results.

Like \citet{inceptionv3}, we used RMSProp for our optimizer, a starting learning rate of 0.045, a learning rate decay every two epochs at an exponential rate of 0.94, and momentum of 0.9. 

Finally, we used the Cleverhans library \cite{papernot2017cleverhans} to implement our adversarial attacks.

\subsubsection{Targeted vs. untargeted attacks}
\citet{obfuscated} state that on ImageNet, accuracy on targeted attacks is a much more meaningful metric to use than accuracy on untargeted attacks. They state that this is because untargeted attacks can cause misclassification of very similar classes (e.g. images of two very similar dog breeds), which is not meaningful.
This is consistent with observations by \citet{kurakin17}.

To that end, as \citet{obfuscated} recommends, all accuracies we report
on ImageNet are for targeted attacks, and all adversarial training was done with targeted attacks.

\subsubsection{Results and discussion}

\textbf{Results with adversarial logit pairing.} We present our main ImageNet results in Tables \ref{fig:wb} and \ref{fig:bb}. All accuracies reported refer to the worst case accuracies among all attacks we tried in each of the two threat models we consider (white box and black box). We used the following attacks in our attack suite, which are all from \citet{madry}, \citet{kurakin17}, and \citet{eat}: Step-LL, Step-Rand, R+Step-LL, R+Step-Rand, Iter-Rand, Iter-LL, PGD-Rand, and PGD-LL. The suffixes "Rand" and "LL" denote targeting a random class and targeting the least likely class, respectively. 
For the multi-step attacks in our suite, we varied the size of the total adversarial perturbation, the size of the perturbation per step, and the number of attack steps. Below are the maximum values of each of these sizes that we tried:

\begin{itemize}
\itemsep0em
  \item Size of total adversarial perturbation: $16/255$ on a scale of 0 to 1
  \item Size of total adversarial perturbation per step: $2/255$ on a scale of 0 to 1
  \item Number of attack steps: $10$
\end{itemize}

All accuracies reported are on the ImageNet validation set.

\begin{table}[h!]
\centering
\begin{tabular}{p{3.5cm}*{2}{c}}
\toprule
\rowcolor{Gray}
 & \textbf{White Box} & \textbf{White Box} \\
\rowcolor{Gray}
\textbf{Method} & \textbf{Top 1} & \textbf{Top 5} \\
\midrule
Regular training & 0.7\% & 4.4 \% \\
\citet{eat} & 1.3\% & 6.5 \% \\
\citet{kurakin17} & 1.5\% & 5.5 \% \\
M-PGD & 3.9\% & 10.3\% \\
ALP & \textbf{27.9\%} & \textbf{55.4\%} \\
\bottomrule
\end{tabular}
  \caption{Comparison of adversarial logit pairing and vanilla adversarial training on ImageNet. All accuracies reported are for \textbf{white box} accuracy on the ImageNet validation set.}
\label{fig:wb}
\end{table}

\begin{table}[h!]
\centering
\begin{tabular}{p{3.5cm}*{2}{c}}
\toprule
\rowcolor{Gray}
 & \textbf{Black Box} & \textbf{Black Box} \\
\rowcolor{Gray}
\textbf{Method} & \textbf{Top 1} & \textbf{Top 5} \\
\midrule
M-PGD & 36.5\% & 62.3\% \\
ALP & 46.7\% & 74.0\% \\
\citet{eat} & \textbf{47.1\%} & \textbf{74.3\%} \\
\bottomrule
\end{tabular}
  \caption{Comparison of adversarial logit pairing and vanilla adversarial training on ImageNet. All accuracies reported are for \textbf{black box} accuracy on the ImageNet validation set.}
\label{fig:bb}
\end{table}

\textbf{Damaging Ensemble Adversarial Training.} 
Ensemble adversarial training \cite{eat} reported the state of the art for ImageNet black box attacks at 66.6\% Top-1 black box accuracy for InceptionV3. Here, we present a black box attack that significantly damages the defense proposed in Ensemble Adversarial Training. 

We construct a black box attack by taking an ALP-trained ImageNet model and constructing a transfer attack with that model. Out of all of the attacks we tried, we found that the Iter-Rand attack \cite{kurakin17} was the strongest against Ensemble Adversarial Training. This attack reduces the accuracy of Ensemble Adversarial Training from 66.6\% Top-1 black box accuracy to \textbf{47.1\%}. 

We hypothesize that the reason this attack was so strong is because it came from a model that had used multi-step adversarial training. The attacks used in \citet{eat} all came from models that had been trained with one or two steps of adversarial training. Black box results from \citet{madry} generally show that examples from adversarially trained models are more likely to transfer to other models.

Thus, we recommend adversarial training with full iterative attacks to provide a minimal level of white box and black box robustness on ImageNet. When testing black box accuracy on ImageNet, we recommend using attacks from models that have been adversarially trained with multiple steps to get a sense of the strongest possible black box attack. Adversarial training with one step attacks (even with ensemble training) on ImageNet can be broken in both the white box and black box case.

 \textbf{Discussion.} Firstly, our results show that PGD adversarial training \textit{can} lead to convergence on ImageNet when combined with synchronous gradient updates and large batch sizes. Scaling adversarial training to ImageNet had not been previously shown before and had been an open question \cite{obfuscated}. Multi-step adversarial training does show an improvement on white box accuracies from the previous state-of-the-art, from 1.5\% to \textbf{3.9\%}.
 
  Secondly, we see that ALP further improves white box accuracy from the adversarial training baseline -- showing an improvement from 3.9\% to \textbf{27.9\%}. Adversarial logit pairing also improves black box accuracy from the M-PGD baseline, going from 36.5\% to \textbf{47.1\%}.
  
  Finally, these results show that adversarial logit pairing achieves \textbf{state of the art on ImageNet} on white box attacks -- with a drastic 20x improvement over the previous state of the art \cite{kurakin17, eat}. We do this while still matching the black box results of Ensemble Adversarial Training, the current state-of-the-art black box defense \cite{eat}. 

We hypothesize that adversarial logit pairing works well because it provides an additional
prior that regularizes the model toward a more accurate understanding of the classes.
If we train the model with only the cross-entropy loss, it is prone to learning spurious
functions that fit the training distribution but have undefined behavior off the training
manifold.
Adversarial training adds additional information about the structure of the space. By
adding an assumption that small perturbations should not change the class, regardless of
direction, adversarial training introduces another prior that forces the model to select
functions that have sensible behavior over a much larger region.
However, adversarial training does not include any information about the relationship
between a clean adversarial example and the adversarial version of the same example.
In adversarial training, we might take an image of a cat, perturb it so the model thinks
it is a dog, and then ask the model to still recognize the image as a cat.
There is no signal to tell the model that the adversarial example is similar specifically
to the individual cat image that started the process.
Adversarial logit pairing forces the explanations of a clean example and the corresponding
adversarial example to be similar.
This is essentially a prior encouraging the model to learn logits that are a function of
the truly meaningful features in the image (position of cat ears, etc.)
and ignore the features that are spurious (off-manifold directions introduced by adversarial
perturbations).
We can also think of the process as distilling \citep{hinton2015distilling} the knowledge from the clean domain into
the adversarial domain and vice versa.

Similar to the dip in clean accuracy on CIFAR-10 reported by \citet{madry}, we found that our models have a slight dip in clean accuracy to 72\%. However, we believe this is outweighed by the large gains in adversarial accuracies.

\subsubsection{Comparison of different architectures}
Model architecture plays a role in adversarial robustness \cite{dogus}, and models with higher capacities tend to be more robust \cite{madry,kurakin17}. Since ImageNet is a particularly challenging dataset, we think that studying different model architectures in conjunction with adversarial training would be valuable. In this work, we primarily studied InceptionV3 to offer better comparisons to previous literature. With the rest of our available computational resources, we were able to study an additional model (ResNet-101) to see if residual connections impacted adversarial robustness. We used ALP to train the models, and results are reported in Tables \ref{fig:wb_resnet_101} and \ref{fig:bb_resnet_101}.

\begin{table}[h!]
\centering
\begin{tabular}{*{5}{c}}
\toprule
\rowcolor{Gray}
\textbf{Method} & \textbf{White Box Top 1} & \textbf{White Box Top 5} \\
\midrule
InceptionV3 & 27.9\% & 55.4\% \\
ResNet-101 & 30.2\% & 55.8\% \\
\bottomrule
\end{tabular}
  \caption{Comparison of InceptionV3 and ResNet101 on ImageNet. All accuracies reported are for \textbf{white box} accuracy on the ImageNet validation set.}
\label{fig:wb_resnet_101}
\end{table}

\begin{table}[h!]
\centering
\begin{tabular}{*{5}{c}}
\toprule
\rowcolor{Gray}
\textbf{Method} & \textbf{Black Box Top 1} & \textbf{Black Box Top 5} \\
\midrule
InceptionV3 & 46.7\% & 74.0\% \\
ResNet-101 & 36.0\% & 62.2\% \\
\bottomrule
\end{tabular}
  \caption{Comparison of InceptionV3 and ResNet101 on ImageNet. All accuracies reported are for \textbf{black box} accuracy on the ImageNet validation set.}
\label{fig:bb_resnet_101}
\end{table}

\subsection{Clean logit pairing results}

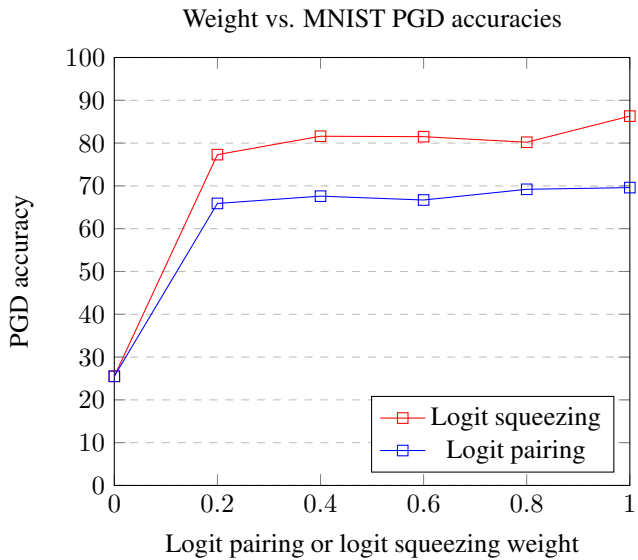
\begin{figure}[!h]
\centering
\begin{tikzpicture}
\begin{axis}[
    title={Weight vs. MNIST PGD accuracies},
    xlabel={Logit pairing or logit squeezing weight},
    ylabel={PGD accuracy},
    xmin=0, xmax=1,
    ymin=0, ymax=100,
    xtick={0, 0.2, 0.4, 0.6, 0.8, 1.0},
    ytick={0, 10, 20, 30, 40, 50, 60, 70, 80, 90, 100},
    legend pos=south east,
    ymajorgrids=true,
    grid style=dashed,
]
 
\addplot[
    color=red,
    mark=square,
    ]
    coordinates {
    (0, 25.5)(0.2, 77.3)(0.4, 81.6)(0.6, 81.5)(0.8, 80.2)(1.0, 86.3)
    };

\addplot[
    color=blue,
    mark=square,
    ]
    coordinates {
    (0, 25.5)(0.2, 65.9)(0.4, 67.6)(0.6, 66.7)(0.8, 69.2)(1.0, 69.6)
    };

\legend{Logit squeezing, Logit pairing}
 
\end{axis}
\end{tikzpicture}
\caption{Varying the logit pairing weight for MNIST}
\label{fig:mnist_lp_weight}
\end{figure}

We experimented with clean logit pairing on MNIST, and we found that it gave surprisingly high results on white box and black box accuracies. As mentioned in our methods section, we augmented images with Gaussian noise first and then applied clean logit pairing or logit squeezing. Logit squeezing resulted in slightly higher PGD accuracies than CLP (detailed in Figure \ref{fig:mnist_lp_weight}). Table \ref{fig:mnist_clp} contains our final MNIST results on clean logit squeezing. For evaluation with PGD, we used the same attack parameters as our evaluation for adversarial logit pairing.

\begin{table}[h!]
\centering
\begin{tabular}{*{5}{c}}
\toprule
\rowcolor{Gray}
\textbf{Method} & \textbf{White box} & \textbf{Black box} & \textbf{Clean} \\
\midrule
M-PGD & 93.2\% & 96.0\% & 98.8\% \\
Logit squeezing & 86.4\% & 96.8\% & 99.0\% \\

\bottomrule
\end{tabular}
  \caption{Comparison of clean logit squeezing and vanilla adversarial training on MNIST. All accuracies reported are for the PGD attack.}
\label{fig:mnist_clp}
\end{table}

As Table \ref{fig:mnist_clp} shows, clean logit squeezing is competitive with adversarial training, despite the large reduction in computational cost.

We also experimented with changing the weight of logit pairing and logit squeezing to see if it acts as a controllable parameter, and results are in Figure \ref{fig:mnist_lp_weight}.

One thing to note about Figure \ref{fig:mnist_lp_weight} is that simply augmenting images with Gaussian noise is enough to bring up PGD accuracy to around 25 \% -- about 2.5 times better than guessing at random. We would like to emphasize that the noise was added during training time, not test time. Noise and other randomized test time defenses have been shown to be broken by \citet{obfuscated}. Going from nearly 0 percent PGD accuracy to 25 percent with just Gaussian noise suggests that there could be other simple changes to training procedures that result in better robustness against attacks.

The below table reports results on SVHN with the PGD attack. Below are the attack parameters used:

\begin{itemize}
  \item Size of total adversarial perturbation: $12/255$ on a scale of 0 to 1
  \item Size of total adversarial perturbation per step: $3/255$ on a scale of 0 to 1
  \item Number of attack steps: $10$
\end{itemize}

\begin{table}[h!]
\centering
\begin{tabular}{*{5}{c}}
\toprule
\rowcolor{Gray}
\textbf{Method} & \textbf{White Box} & \textbf{Black Box} & \textbf{Clean} \\
\midrule
M-PGD & 44.4\% & 55.4\% & 96.9\% \\
CLP & 39.1\% & 55.8\% & 95.5\% \\
\bottomrule
\end{tabular}
\caption{Clean logit pairing results on SVHN.}
\label{table:top5}
\end{table}

As the above tables show, clean logit pairing is competitive with adversarial training for black box results, \textit{despite the large reduction in computational cost}. Adversarial training with multi-step attacks scales with the number of steps per attack because full backpropagation is completed with each attack step. In other words, if the normal training time of a model is N, adversarial training with k steps per attack will roughly cause the full training time of the model to be kN. 

In contrast, the cost of CLP in terms of floating point operations and memory
consumption is $O(1)$ in the sense that it does not scale
with the number or size of hidden layers in the model,
input image size, or number of attack steps. 
It does scale with the number of logits, but this is negligible compared
to the other factors. Typically the number of logits is determined by the task
(10 for CIFAR-10, 100 for ImageNet) and remains fixed, while 
other factors like model size are desirable to increase. For example, binary classification is a common task in many real world applications like spam and fraud detection.

We hope that CLP points the way to further effective defenses
that are essentially free.
Defenses with low computational cost are more likely to be
adopted since they require fewer resources.
The future of machine learning security is much brighter if security
can be accomplished without a major tradeoff against training
efficiency.
 
\section{Comparison to other possible approaches}

Logit pairing is similar to two other approaches that have been previously shown
to improve adversarial robustness:
\newterm{label smoothing} and \newterm{mixup}.

Label smoothing \citep{inceptionv3} consists of training a
classifier using soft targets for the cross-entropy loss
rather than hard targets.
The correct class is given a target probability of $1-\delta$
and the remaining $\delta$ probability mass is divided uniformly
between the incorrect classes.
This technique is somewhat related to our work because smaller
logits will generally cause smoother output distributions,
but note that label smoothing would be satisfied to have
very large logits so long as the probabilities after normalization
are smooth.
\citet{WardeFarley16} showed that label smoothing offers a small
amount of robustness to adversarial examples, and it is
included by default in the CleverHans tutorial on adversarial
examples \citep{papernot2017cleverhans}.

Mixup \citep{mixup} trains the model on input points that
are interpolated between training examples. At these
interpolated input points, the output target is formed
by similarly interpolating between the target distributions
for each of the training examples.
\citet{mixup} reports that mixup increases robustness to
adversarial examples.

We present our results comparing adversarial logit pairing to label smoothing and mixup in Table \ref{table:compare}. Here, we use ResNet-101 on ImageNet, and all evaluations are with PGD. We find that adversarial logit pairing provides a much stronger defense than either of these two approaches.

\begin{table}[h!]
\centering
\begin{tabular}{*{5}{c}}
\toprule
\rowcolor{Gray}
\textbf{Method} & \textbf{Top 1} & \textbf{Top 5} \\
\midrule
Mixup & 0.1\% & 1.5\% \\
Label smoothing & 1.6\% & 10.0\% \\
ALP & 30.2\% & 55.8\% \\
\bottomrule
\end{tabular}
\caption{White box accuracies under \citet{madry}
attack on ImageNet for label smoothing, mixup, and adversarial logit pairing.}
\label{table:compare}
\end{table}

Besides these related methods of defense against adversarial examples, ALP is also similar to a method of
{\em semi-supervised learning}: virtual adversarial training \citep{miyato2017virtual}.
Virtual adversarial training (VAT) is a method designed to learn from {\em unlabeled data}
by training the model to resist adversarial perturbations of unlabeled data.
The goal of VAT is to reduce test error when training with a small set of labeled
examples, not to cause robustness to adversarial examples.
VAT consists of:
\begin{enumerate}
    \item Construct adversarial examples by perturbing unlabeled examples
    \item Specifically, make the adversarial examples by maximizing the KL divergence between the predictions on the clean examples and the predictions on the adversarial examples.
    \item During model training, add a loss term that minimizes KL divergence between predictions on clean and adversarial examples.
\end{enumerate}
ALP does not include (1) or (2) but does resemble (3).
Both ALP and VAT encourage the full distribution of predictions on clean and adversarial
examples to be similar. VAT does so using a non-symmetric loss applied to the output
probabilities; ALP does so using a symmetric loss applied to the logits.
During the design of our defense, we found that VAT offered an improvement over the baseline
Madry model on MNIST, but ALP consistently performed better than VAT on MNIST across
several hyperparameter values. ALP also performed better than VAT with the direction of the KL
flipped.
We therefore focused on further developing ALP.
The better performance of ALP than VAT may be due to the fact that the KL divergence can suffer
from saturating gradients or it may be due to the fact that the KL divergence is invariant to
a shift of all the logits for an individual example while the logit pairing loss is not.
Logit pairing encourages the logits for the clean and adversarial example to be centered
on the same mean logit value, which doesn't change the information in the output probabilities
but may affect the learning dynamics.

\section{Conclusion and Future Work}
In conclusion, we implement adversarial training at unprecendented scale and present logit pairing as a defense. The experiments in this paper were run on NVIDIA p100s, but with the recent availability of much more powerful hardware (NVIDIA v100s, Cloud TPUs, etc.), we believe that defenses for adversarial examples on ImageNet will become even more scalable.
Specifically our contributions are:
\begin{itemize}
    \item We answer the open question as to whether adversarial training scales to ImageNet. 
    \item We introduce adversarial logit pairing (ALP), an extension to adversarial training that greatly increases its effectiveness.
    \item We introduce clean logit pairing and logit squeezing, low-cost alternatives to adversarial training that can increase the adoption of robust machine learning due to their requirement of very few resources.
    \item We demonstrate that ALP-trained models can generate attacks strong enough to significantly damage the previously state of the art Ensemble Adversarial Training defense, which was used
    by all 10 of the top defense teams in the NIPS 2017 competition on adversarial examples.
    \item We show that adversarial logit pairing achieves the \textbf{state of the art} defense for white box and black box attacks on ImageNet.
\end{itemize}

Our results suggest that feature pairing (matching adversarial and clean intermediate features instead of logits)
may also prove useful in the future.

One limitation to our defenses is that they are not
currently certified or verified (there is no proof that the
true robustness of the system is similar to the robustness
that we measured empirically).
Research into certification and verification methods \citep{katz2017reluplex,kolter2017provable,raghunathan2018certified,sinha2018certifiable}
could  make it possible to certify or verify these same networks in future
work. Current certification methods do not scale to the size of models we trained here
or are only able to provide tight certification bounds for models that were
trained to be easy to certify using a specific certification method.

We would like to note that these defense mechanisms are not
yet sufficient to secure machine learning in a real system
(see many of the concerns raised by
\cite{brown2017adversarial} and
 \citet{adv_spheres}),
and that attacks could be developed against our work in the future. Here, we use ALP in conjunction with the PGD attack since it is the strongest attack presented so far, but since ALP is independent of the actual attack it is used with, it is conceivable that ALP could be used in conjunction with future attacks to develop stronger defenses.
In conclusion, we present our defense as the current state of the art
of research into defenses, and we believe it will serve
as one step along the path to a complete defense in the future.

\ifdefined\isaccepted{
\section*{Acknowledgements}
We thank Tom Brown for helpful feedback on drafts of this article.
}\fi

\bibliography{example_paper}

\begin{thebibliography}{25}
\providecommand{\natexlab}[1]{#1}
\providecommand{\url}[1]{\texttt{#1}}
\expandafter\ifx\csname urlstyle\endcsname\relax
  \providecommand{\doi}[1]{doi: #1}\else
  \providecommand{\doi}{doi: \begingroup \urlstyle{rm}\Url}\fi

\bibitem[Aditi~Raghunathan(2018)]{raghunathan2018certified}
Aditi~Raghunathan, Jacob~Steinhardt, Percy~Liang.
\newblock Certified defenses against adversarial examples.
\newblock \emph{International Conference on Learning Representations}, 2018.
\newblock URL \url{https://openreview.net/forum?id=Bys4ob-Rb}.

\bibitem[Aman~Sinha(2018)]{sinha2018certifiable}
Aman~Sinha, Hongseok~Namkoong, John~Duchi.
\newblock Certifiable distributional robustness with principled adversarial
  training.
\newblock \emph{International Conference on Learning Representations}, 2018.
\newblock URL \url{https://openreview.net/forum?id=Hk6kPgZA-}.

\bibitem[Athalye et~al.(2018)Athalye, Carlini, and Wagner]{obfuscated}
Athalye, Anish, Carlini, Nicholas, and Wagner, David.
\newblock Obfuscated gradients give a false sense of security: Circumventing
  defenses to adversarial examples.
\newblock Technical report, arXiv, 2018.
\newblock URL \url{https://arxiv.org/abs/1802.00420}.

\bibitem[Brown et~al.(2017)Brown, Man{\'e}, Roy, Abadi, and
  Gilmer]{brown2017adversarial}
Brown, Tom~B, Man{\'e}, Dandelion, Roy, Aurko, Abadi, Mart{\'\i}n, and Gilmer,
  Justin.
\newblock Adversarial patch.
\newblock \emph{arXiv preprint arXiv:1712.09665}, 2017.

\bibitem[Buckman et~al.(2018)Buckman, Roy, Raffel, and
  Goodfellow]{buckman2018thermometer}
Buckman, Jacob, Roy, Aurko, Raffel, Colin, and Goodfellow, Ian.
\newblock Thermometer encoding: One hot way to resist adversarial examples.
\newblock \emph{ICLR}, 2018.
\newblock URL \url{https://openreview.net/forum?id=S18Su--CW}.

\bibitem[Cubuk et~al.(2017)Cubuk, Zoph, Schoenholz, and Le]{dogus}
Cubuk, E.~D., Zoph, B., Schoenholz, S., and Le, Q.
\newblock Intriguing properties of adversarial examples.
\newblock Technical report, arXiv, 2017.
\newblock URL \url{https://arxiv.org/pdf/1711.02846.pdf}.

\bibitem[Gilmer et~al.(2018)Gilmer, Metz, Faghri, Schoenholz, Raghu,
  Wattenberg, and Goodfellow]{adv_spheres}
Gilmer, Justin, Metz, Luke, Faghri, Fartash, Schoenholz, Samuel~S., Raghu,
  Maithra, Wattenberg, Martin, and Goodfellow, Ian.
\newblock Adversarial spheres.
\newblock In \emph{ICLR 2018 workshop}, 2018.
\newblock URL \url{https://arxiv.org/abs/1801.02774}.

\bibitem[Gomez et~al.(2017)Gomez, Ren, Urtasun, and Grosse]{revnets}
Gomez, Aidan, Ren, Mengye, Urtasun, Raquel, and Grosse, Roger.
\newblock The reversible residual network: Backpropagation without storing
  activations.
\newblock In \emph{NIPS 2017}, 2017.
\newblock URL \url{https://arxiv.org/abs/1707.04585}.

\bibitem[Goodfellow et~al.(2014)Goodfellow, Shlens, and
  Szegedy]{goodfellow2014explaining}
Goodfellow, Ian~J, Shlens, Jonathon, and Szegedy, Christian.
\newblock Explaining and harnessing adversarial examples.
\newblock \emph{arXiv preprint arXiv:1412.6572}, 2014.

\bibitem[Hinton et~al.(2015)Hinton, Vinyals, and Dean]{hinton2015distilling}
Hinton, Geoffrey, Vinyals, Oriol, and Dean, Jeff.
\newblock Distilling the knowledge in a neural network.
\newblock \emph{arXiv preprint arXiv:1503.02531}, 2015.

\bibitem[Katz et~al.(2017)Katz, Barrett, Dill, Julian, and
  Kochenderfer]{katz2017reluplex}
Katz, Guy, Barrett, Clark, Dill, David~L, Julian, Kyle, and Kochenderfer,
  Mykel~J.
\newblock Reluplex: An efficient smt solver for verifying deep neural networks.
\newblock In \emph{International Conference on Computer Aided Verification},
  pp.\  97--117. Springer, 2017.

\bibitem[Kolter \& Wong(2017)Kolter and Wong]{kolter2017provable}
Kolter, J~Zico and Wong, Eric.
\newblock Provable defenses against adversarial examples via the convex outer
  adversarial polytope.
\newblock \emph{arXiv preprint arXiv:1711.00851}, 2017.

\bibitem[Kurakin et~al.(2017{\natexlab{a}})Kurakin, Goodfellow, and
  Bengio]{kurakin17}
Kurakin, Alexey, Goodfellow, Ian, and Bengio, Samy.
\newblock Adversarial machine learning at scale.
\newblock In \emph{ICLR 2017}, 2017{\natexlab{a}}.
\newblock URL \url{https://arxiv.org/abs/1611.01236}.

\bibitem[Kurakin et~al.(2017{\natexlab{b}})Kurakin, Goodfellow, and
  Bengio]{kurakin17physical}
Kurakin, Alexey, Goodfellow, Ian, and Bengio, Samy.
\newblock Adversarial examples in the physical world.
\newblock In \emph{ICLR'2017 Workshop}, 2017{\natexlab{b}}.
\newblock URL \url{https://arxiv.org/abs/1607.02533}.

\bibitem[Madry et~al.(2017)Madry, Makelov, Schmidt, Tsipras, and Vladu]{madry}
Madry, A., Makelov, A., Schmidt, L., Tsipras, D., and Vladu, A.
\newblock Towards deep learning models resistant to adversarial attacks.
\newblock Technical report, arXiv, 2017.
\newblock URL \url{https://arxiv.org/pdf/1706.06083.pdf}.

\bibitem[Miyato et~al.(2017)Miyato, Maeda, Koyama, and
  Ishii]{miyato2017virtual}
Miyato, Takeru, Maeda, Shin-ichi, Koyama, Masanori, and Ishii, Shin.
\newblock Virtual adversarial training: a regularization method for supervised
  and semi-supervised learning.
\newblock \emph{arXiv preprint arXiv:1704.03976}, 2017.

\bibitem[Nicolas~Papernot(2017)]{papernot2017cleverhans}
Nicolas~Papernot, Nicholas~Carlini, Ian Goodfellow Reuben Feinman Fartash
  Faghri Alexander Matyasko Karen Hambardzumyan Yi-Lin Juang Alexey Kurakin
  Ryan Sheatsley Abhibhav Garg Yen-Chen~Lin.
\newblock cleverhans v2.0.0: an adversarial machine learning library.
\newblock \emph{arXiv preprint arXiv:1610.00768}, 2017.

\bibitem[Papernot et~al.(2016)Papernot, McDaniel, Wu, Jha, and
  Swami]{papernot2016distillation}
Papernot, Nicolas, McDaniel, Patrick, Wu, Xi, Jha, Somesh, and Swami,
  Ananthram.
\newblock Distillation as a defense to adversarial perturbations against deep
  neural networks.
\newblock In \emph{Security and Privacy (SP), 2016 IEEE Symposium on}, pp.\
  582--597. IEEE, 2016.

\bibitem[Sietsma \& Dow(1991)Sietsma and Dow]{SietsmaDow91}
Sietsma, J. and Dow, R.
\newblock Creating artificial neural networks that generalize.
\newblock \emph{Neural Networks}, 4\penalty0 (1):\penalty0 67--79, 1991.

\bibitem[Szegedy et~al.(2013)Szegedy, Zaremba, Sutskever, Bruna, Erhan,
  Goodfellow, and Fergus]{szegedy2013intriguing}
Szegedy, Christian, Zaremba, Wojciech, Sutskever, Ilya, Bruna, Joan, Erhan,
  Dumitru, Goodfellow, Ian, and Fergus, Rob.
\newblock Intriguing properties of neural networks.
\newblock \emph{arXiv preprint arXiv:1312.6199}, 2013.

\bibitem[Szegedy et~al.(2016)Szegedy, Vanhoucke, Ioffe, Shlens, and
  Wojna]{inceptionv3}
Szegedy, Christian, Vanhoucke, Vincent, Ioffe, Sergey, Shlens, Jon, and Wojna,
  Zbigniew.
\newblock Rethinking the inception architecture for computer vision.
\newblock In \emph{Proceedings of the IEEE Conference on Computer Vision and
  Pattern Recognition}, pp.\  2818--2826, 2016.

\bibitem[{Tram{\`e}r} et~al.(2018){Tram{\`e}r}, {Kurakin}, {Papernot}, {Boneh},
  and {McDaniel}]{eat}
{Tram{\`e}r}, F., {Kurakin}, A., {Papernot}, N., {Boneh}, D., and {McDaniel},
  P.
\newblock Ensemble adversarial training: Attacks and defenses.
\newblock In \emph{ICLR 2018}, 2018.
\newblock URL \url{https://arxiv.org/abs/1705.07204}.

\bibitem[Warde-Farley \& Goodfellow(2016)Warde-Farley and
  Goodfellow]{WardeFarley16}
Warde-Farley, David and Goodfellow, Ian.
\newblock Adversarial perturbation of deep neural networks.
\newblock In Hazan, Tamir, Papandreou, George, and Tarlow, Daniel (eds.),
  \emph{Perturbation, Optimization, and Statistics}. MIT Press, 2016.

\bibitem[Xu et~al.(2017)Xu, Evans, and Qi]{xu2017feature}
Xu, Weilin, Evans, David, and Qi, Yanjun.
\newblock Feature squeezing: Detecting adversarial examples in deep neural
  networks.
\newblock \emph{arXiv preprint arXiv:1704.01155}, 2017.

\bibitem[Zhang et~al.(2017)Zhang, Cisse, Dauphin, and Lopez-Paz]{mixup}
Zhang, H., Cisse, M., Dauphin, Y., and Lopez-Paz, D.
\newblock Mixup: Beyond empirical risk minimization.
\newblock Technical report, arXiv, 2017.
\newblock URL \url{https://arxiv.org/pdf/1710.09412.pdf}.

\end{thebibliography}
\bibliographystyle{icml2018}

\end{document}